\documentclass[letterpaper, 10 pt, conference]{ieeeconf}  

\IEEEoverridecommandlockouts                              

\usepackage[bookmarks=true]{hyperref}
\usepackage{color}
\usepackage{cite}
\usepackage{graphicx}
\usepackage{bm}
\usepackage{tabularx}
\usepackage{floatrow}
\usepackage[utf8]{inputenc}
\usepackage{amsmath}
\usepackage[utf8]{inputenc} 
\usepackage[T1]{fontenc}    
\usepackage[mathscr]{eucal}
\usepackage{amsfonts}
\usepackage{dblfloatfix} 
\usepackage{epsfig}
\usepackage{makecell}
\usepackage{multirow}
\usepackage{hyperref}
\usepackage{dsfont}
\usepackage{breqn}
\usepackage{amssymb}
\usepackage{amsmath,amsthm}
\usepackage{amsfonts}       
\usepackage{nicefrac}       
\newtheorem{theorem}{Theorem}
\usepackage{amsthm}

\newtheorem{problem}{Problem}

\usepackage{mathtools,algorithm}
\usepackage[noend]{algpseudocode}
\title{\LARGE \bf
Graph Policy Gradients for Large Scale Unlabeled Motion Planning with Constraints
}

\author{Arbaaz Khan$^{1}$, Vijay Kumar$^{1}$, Alejandro Ribeiro$^{1}$ 
\thanks{$^{1}$Authors are with GRASP Lab, University of Pennsylvania, USA
        {\tt\small arbaazk@seas.upenn.edu}}%
}

\begin{document}

\maketitle
\thispagestyle{empty}
\pagestyle{empty}

\begin{abstract}
In this paper, we present a learning method to solve the unlabelled motion problem with motion constraints and space constraints in 2D space for a large number of robots. To solve the problem of arbitrary dynamics and constraints we propose formulating the problem as a multi-agent problem. In contrast to previous works that propose using learning solutions for unlabelled motion planning with constraints, we are able to demonstrate the scalability of our methods for a large number of robots. The curse of dimensionality one encounters when working with a large number of robots is mitigated by employing a graph convolutional neural (GCN) network to parametrize policies for the robots. The GCN reduces the dimensionality of the problem by learning filters that aggregate information among robots locally, similar to how a convolutional neural network is able to learn local features in an image. Additionally, by employing a GCN we are also able to overcome the computational overhead of training policies for a large number of robots by first training graph filters for a small number of robots followed by zero-shot policy transfer to a larger number of robots. We demonstrate the effectiveness of our framework through various simulations. A video describing our results can be found \href{https://youtu.be/VzqLNhvBJxk}{here}. 
\end{abstract}

\section{Introduction}
\label{sec:Sec1}
In robotics, one is often tasked with designing algorithms to coordinate teams of robots to achieve a task. Some examples of such tasks are formation flying \cite{desai2001modeling,khan2019graph,alonso2015multi}, perimeter defense \cite{shishika2018local}, surveillance \cite{saldana2016dynamic}. In this paper, we concern ourselves with scenarios where a team of homogeneous or identical robots must execute a set of identical tasks such that each robot executes only one task, but it does not matter which robot executes which task. Concretely, this paper studies the concurrent goal assignment and trajectory planning problem where robots must simultaneously assign goals and plan motion primitives to reach assigned goals. This is the unlabelled multi-robot planning problem where one must simultaneously solve for goal assignment and trajectory optimization.

In the past, several methods have been proposed to achieve polynomial-time solutions for the unlabelled motion planning problem \cite{adler2015efficient, macalpine2015scram,yu2012distance,turpin2014capt}. The common theme among these methods is the design of a heuristic best suited for the robots and the environment. For example \cite{turpin2014capt} minimizes straight line distances and solves for the optimal assignment using the Hungarian algorithm.
However, when additional constraints such as constraints on dynamics, presence of obstacles in the space, desired goal orientations solving for a simple heuristic is no longer the optimal solution. In fact, one can think of the robot to goals matching problem with constraints as an instance of the minimum cost perfect matching problem with conflict pair constraints (MCPMPC) which is in fact known to be strongly $\mathcal{NP}$-hard \cite{darmann2011paths}.
\begin{figure}[t!]
  \centering
  \includegraphics[scale=0.40]{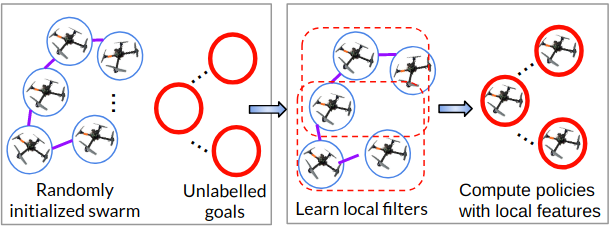}
  \caption{\textbf{Graph Policy Gradients for Unlabeled Motion Planning.} $\mathbf{N}$ Robots and $\mathbf{N}$ goals are randomly initialized in some space. To define the graph, each robot is a node and threshold is connected to $\mathbf{n}$ nearest robots. Each robot observes relative positions of $\mathbf{m}$ nearest goals and other entities within a sensing region. Information from K-hop neighbors is aggregated at each node by learning local filters. Local features along with the robots own features are then used to learn policies to produce desired coverage behavior. \label{fig:mainfig}}
\end{figure}
In contrast to previous literature that relies on carefully designed but ultimately brittle heuristic functions, we hypothesize using reinforcement learning (RL) to compute an approximate solution for the assignment and planning problem without relaxing any constraints. When using RL to learn policies for the robots, it might be possible to simply use a cost function that is independent of any dynamics or constraints, i.e a cost function that is set to a high value if all goals are covered and zero otherwise. The idea of casting the unlabelled motion planning problem as a multi-agent reinforcement learning (MARL) problem has been explored before by Khan et. al \cite{khan2019learning}. They propose learning individual policies for each robot which are trained by using a central Q-network that coordinates information among all robots. However, the authors of \cite{khan2019learning} themselves note that the key drawback of their method is that does not scale as the number of robots increase. Further, there is a huge computational overhead attached with training individual policies for each robot. 

There exist two key obstacles in learning policies that scale with the number of robots: increase in dimensionality and partial observability. Consider an environment with $\mathbf{N}$ robots (This paper uses bold font to denote collection of items, vectors and matrices). In a truly decentralized setting, each robot can at best only partially sense its environment. In order for a robot to learn a meaningful control policy, it must communicate with some subset of all agents, $\mathbf{n} \subseteq \mathbf{N}$. Finding the right subset of neighbors to learn from  which is in itself a research problem. 
Additionally, to ensure scalability, as the number of robots increase, one needs to ensure that the cardinality of the subset of neighbors that each robot must interact with $|\mathbf{n}|$, remains fixed or grows very slowly.

To achieve scalable multi-robot motion planning, we look to exploit the inherent graph structure among the robots and learn policies using local features only. We hypothesize that graph convolutional neural networks (GCNs) \cite{kipf2016semi,wu2019comprehensive} can be a good candidate to parametrize policies for robots as opposed to the fully connected networks used in \cite{khan2019learning}. GCNs work similar to convolutional neural networks (CNNs) and can be seen as an extension of CNNs to data placed on irregular graphs instead of a two dimensional grid such as an image. A brief overview of the workings of a GCN is provided in Sec \ref{sec:GCNs}. 

We define a graph $\mathcal{G} = (\mathbf{V},\mathbf{E})$  where $\mathbf{V}$ is the set of nodes representing the robots and $\mathbf{E}$ is the set of edges defining relationships between them. These relationships can be arbitrary and are user defined. For example in this work we define edges between robots based on the Euclidean distance between them. This graph acts as a support for the data vector $\textbf{x}=[\mathbf{x}_1,\ldots,\mathbf{x}_N]^\top$ where $\mathbf{x}_n$ is the state representation of robot $n$. The GCN consists of multiple layers of graph filters and at each layer the graph filters extract local information from a node's neighbors, similar to how CNNs learn filters that extract features in a neighborhood of pixels. The information is propagated forward between layers after passing it through a non linear activation function similar to how one would propagate information in a CNN. The output of the final layer is given by $\Pi = [\pi_1,\ldots,\pi_N]$,  where $\pi_1,\ldots,\pi_N$ are independent control policies for the robots. During training, each robot rolls out its own trajectory by executing its respective policy. Each robot also collects a centralized reward and using policy gradient methods ~\cite{sutton1998reinforcement} the weights of the GCN are updated. Further, since the robots are homogeneous and the graph filters only learn local information, we circumvent the computational burden of training many robots by training the GCN on only a small number of robots but during inference use the same filter across all robots. 

We call this algorithm Graph Policy Gradients (GPG) (Fig.\ref{fig:mainfig}) and first proposed it in our earlier work \cite{khan2019graph} where it was used to control swarms of robots with designated goals. This is in contrast to this paper which focuses on using GPG for simultaneous goal assignment and trajectory optimization. Through varied simulations we demonstrate that GPG provides a suitable solution that extends the learning formulation of the unlabeled motion planning algorithm to have the ability to scale to many robots. 
\section{Problem Formulation}
\label{sec:problem_formulation}
Consider a Euclidean space $\mathbb{R}^2$ populated with $\mathbf{N}$ homogeneous disk-shaped robots and $\mathbf{N}$ goals of radius $\text{R}$. Goal location $\mathbf{g}_m$ is represented as a vector of its position in the XY plane, i.e $\mathbf{g}_m = [x_m,y_m]$. Robot position at time $t$ is denoted as $\mathbf{p}_{nt}$ and is represented as a vector of its position in the XY plane, i.e $\mathbf{p}_{nt} = [x_{nt},y_{nt}]$. Each robot observes its relative position to $\mathbf{n}$ nearest goals where $\mathbf{n}$ is a parameter set by the user and depends on the task. Each robot also observes its relative position to $\mathbf{\hat{n}}$ nearest robots in order to avoid collisions. In the case that obstacles are present, each obstacle is represented by its position in the XY plane $\mathbf{o}_k = (x_k,y_k)$. 
Each robot also observes its relative position to $\mathbf{\tilde{n}}$ nearest obstacles. We motivate this choice of feature representation in Sec.\ref{subsec:permequi}. Thus, the state of robot $n$ at time $t$ is given as:
\begin{equation}
    \mathbf{x}_{nt} := [\mathbf{G}_{\mathbf{n}t},\mathbf{R}_{\mathbf{\hat{n}}t},\mathbf{O}_{\mathbf{\tilde{n}}t}]
\end{equation}
where $\mathbf{G}_{\mathbf{n}t}$ is the vector that gives relative positions of robot $n$ at time $t$ to $\mathbf{n}$ nearest goals, $\mathbf{R}_{\mathbf{\hat{n}}t}$ is the vector that gives relative positions of robot $n$ at time $t$ to $\mathbf{\hat{n}}$ nearest robots and $\mathbf{O}_{\mathbf{\tilde{n}}t}$ is the vector that gives relative positions of robot $n$ at time $t$ to $\mathbf{\tilde{n}}$ nearest obstacles. At each time $t$, each robot executes an action $\mathbf{a}_{nt} \in \mathcal{A}$, that evolves the state of the robot accorwding to some stationary dynamics distribution with conditional density $p(\mathbf{x}_{n t+1}|\mathbf{x}_{nt},\mathbf{a}_{nt})$. In this work, all actions represent continuous control (change in position or change in velocity). Collision avoidance and goal coverage conditions are encoded to yield the desired collision free unlabelled motion planning behavior where all goals are covered. The necessary and sufficient condition to ensure collision avoidance between robots is given as : 
\begin{equation}
\label{eq:collision_avoidance}
    E_c(\mathbf{p}_{it},\mathbf{p}_{jt}) > \delta, \\ \forall i \neq j \in \{1,\ldots \mathbf{N}\}, \forall {t}
\end{equation} 
where $E_c$ is the euclidean distance and $\delta$ is a user-defined minimum separation between robots. The assignment matrix $\phi(t) \in \mathbb{R}^{\mathbf{N} \times \mathbf{N}}$ as 
\begin{equation}
    \phi_{ij}(t) = 
    \begin{cases}
    1, &\text{if }  E_c(\mathbf{p}_i(t),\mathbf{g}_j)  \leq \psi \\
    0, &\text{otherwise}
    \end{cases}
\end{equation}
where $\psi$ is some threshold region of acceptance. The necessary and sufficient condition for all goals to be covered by robots at some time $t=T$ is then:
\begin{equation}
\label{eq:stopping}
    \phi(T)^\top\phi(T) = \textbf{I}_{\mathbf{N}}
\end{equation}
where $\textbf{I}$ is the identity matrix. Lastly, since we are interested in exploiting local symmetry between robots, we define a graph $\mathcal{G}=(\mathbf{V},\mathbf{E})$ where the set of vertices $\mathbf{V}$ represents all the robots. An edge $e \in \mathbf{E}$  is said to exist between two vertices, if:
\begin{equation}
\label{eq:graph_node}
    E_c(\mathbf{p}_{i},\mathbf{p}_{j}) \leq \lambda, \\ \forall i \neq j \in \{1,\ldots \mathbf{N}\}
\end{equation} 
where $\lambda$ is some user defined threshold to connect two robots. Robots cannot externally communicate with any other robots. Robot $n$ can directly communicate only with its one hop neighbors given by $\mathcal{G}$. In order to communicate with robots further away, the communication must be indirect. Robots are given access to this graph and their own state $\mathbf{x}_{nt}$. The problem statement considered in this paper can then be formulated as :
\begin{problem}
\label{prob:prob1}
\textit{Given a set of $\mathbf{N}$ robots with some initial configurations, a set of $\mathbf{N}$ goals a graph $\mathcal{G}$ defining relationships between the robots} $\textbf{g}$\textit{, compute functions $\Pi = [\pi_1,\ldots,\pi_{\mathbf{N}}]$ such that execution actions $\{\mathbf{a}_{1t},\ldots,\mathbf{a}_{\mathbf{N}t}\}= \{\pi_1(\mathbf{a}_{1t}|\mathbf{x}_{1t},\mathcal{G}),\ldots,\pi_{\mathbf{N}}(\mathbf{a}_{\mathbf{N}t}|\mathbf{x}_{\mathbf{N}t},\mathcal{G})\}$ results in a sequence of states for the robots that satisfy Eq.\ref{eq:collision_avoidance} for all time and at some stopping time $t=T$, satisfy the assignment constraint in Eq.\ref{eq:stopping}.}  
\end{problem}
\section{Graph Convolutional Networks}
\label{sec:GCNs}
A graph $\mathcal{G}=({\mathbf{V},\mathbf{E}})$ is described by a set of $\mathbf{N}$ nodes and a set of edges $\mathbf{E} \subseteq \mathbf{V} \times \mathbf{V}$. This graph can be represented by a graph shift operator $\mathbf{S}$, which respects the sparsity of the graph, i.e $s_{ij} = [\mathbf{S}]_{ij}=0$, $\forall$ $i\neq j \text{ and } (i,j) \notin \mathbf{E}$. Adjacency matrices, graph laplacians and their normalized versions are some examples of this graph shift operator which satisfy the sparsity property. At each node, there exists a data signal $\mathbf{x}_n$. Collectively, define $\textbf{x}=[\mathbf{x}_1,\ldots,\mathbf{x}_N]^\top \in \mathbb{R}^{\mathbf{N}}$ as the signal and its support is given by $\mathcal{G}$. 

\begin{figure}[hbt!]
  \centering
  \includegraphics[scale=0.25]{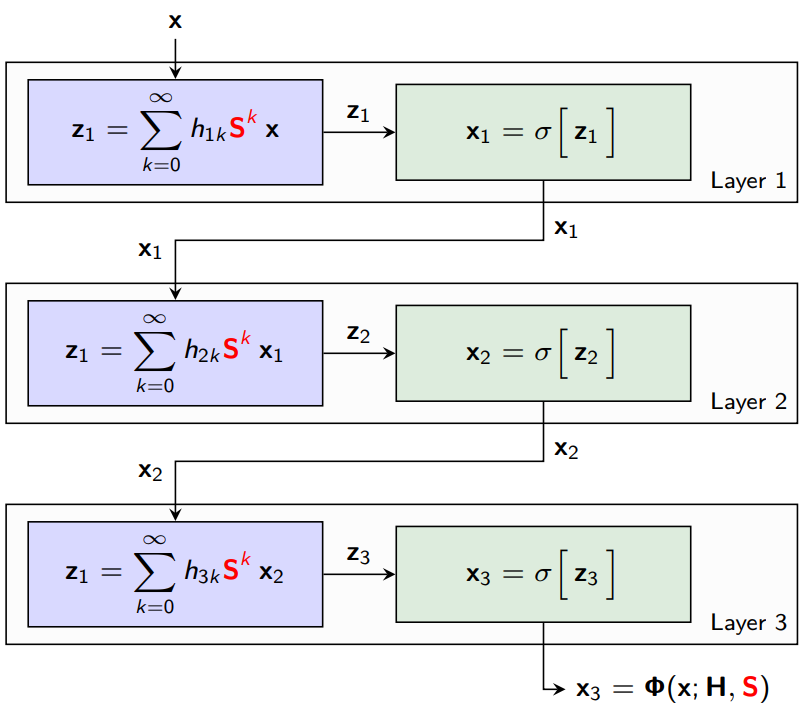}
  \caption{\textbf{Graph Convolutional Networks.} Each layer consists of chosen graph filters with pointwise non linearities.  \label{fig:gnnfig}}
\end{figure}

$\mathbf{S}$ can be used to define a linear map $\mathbf{y}=\mathbf{S}\textbf{x}$ which represents local information exchange between a given node and its neighbors. For example, consider a node $n$ with immediate or one-hop neighbors defined by the set $\mathfrak{B}_n$. The following equation then gives us a simple aggregation of information at node $n$ from its one-hop neighbors $\mathfrak{B}_n$. 
\begin{equation}\label{eq:graph_signal}
    y_n = [\mathbf{S}\mathbf{x}]_n =\sum_{j=n,j\in \mathfrak{B}_n}s_{nj}x_n
\end{equation}
By repeating this operation over all nodes in the graph, one can construct the signal $\mathbf{y}=[y_1,\ldots,y_{\mathbf{N}}]$. Recursive application of this operation yields information from nodes further away i.e the operation $\mathbf{y}^k = \mathbf{S}^k\mathbf{x} = \mathbf{S}(\mathbf{S}^{k-1}\mathbf{x})$ aggregates information from  nodes that are $k$-hops away. Then graph convolutional filters can be defined as polynomials on $\mathbf{S}$ :
\begin{equation}\label{eq:z}
    \mathbf{z} = \sum_{k=0}^{K} h_k \mathbf{S}^k \mathbf{x} = \mathbf{H(S)x}
\end{equation}
Here, the graph convolution is localized to a $K$-hop neighborhood for each node. The output from the graph convolutional filter is fed into a pointwise non-linear activation function $\sigma$ akin to convolutional neural networks.
\begin{equation}
\label{eq:finalapproxform}
    \mathbf{z} = \sigma(\mathbf{H(S)x})
\end{equation}
A graph convolution network can then be composed by stacking several of these layers together as seen in Fig. \ref{fig:gnnfig}
\subsection{Permutation Equivariance}
\label{subsec:permequi}
On-policy RL methods to train policies for robots such as the ones proposed in \cite{khan2019learning} collect trajectories from the current policy and then use these  samples to update the current policy. The initial policy for all robots is a random policy. 
This is the exploration regime in RL. It is highly likely that as one were to randomly explore a space during training, the underlying graph $\mathcal{G}$ that connects robots based on nearest neighbors would change, i.e the graph at $t=0$, say $\mathcal{G}_0$ would be significantly different from the graph at $t=T$, say $\mathcal{G}_T$. We noted in Section \ref{sec:Sec1} that one of the key obstacles with learning scalable policies is that of dimensionality. At a first glance it seems that providing the underlying graph structure would only exacerbate the problem of dimensionality and will result in a large number of graphs that one needs to learn over as the number of robots are increased. 
However, we use a key result from \cite{gama2019stability} that minimizes the number of graphs one must learn over.
The authors of \cite{gama2019stability} show that as long as the topology of the underlying graph is fixed, the output of the graph convolution remains the same even under node reordering. More concretely if given a set of permutation matrices $\bm{\mathcal{P}} = \{\mathbf{{P}}\in \{0,1\}^{\mathbf{N} \times \mathbf{N}}\ : \mathbf{{P1=1}},\mathbf{{P}^\top1 =1}\}$ such that the operation $\mathbf{\Bar{P}x}$ permutes elements of the vector $\mathbf{x}$, then it can be shown that 
\begin{theorem}\label{theorem:permutationequivariance}
Given a graph $\mathcal{G}=(\mathbf{V},\mathbf{E})$ defined with a graph shift operator $\mathbf{S}$ and $\hat{\mathcal{G}}$ to be the  permuted graph with $\mathbf{\hat{S}} = \mathbf{{P}}^{\top} \mathbf{S} \mathbf{{P}}$ for $\mathbf{{P}} \in \bm{\mathcal{P}}$ and any $\mathbf{x} \in \mathbb{R}^{\mathbf{N}}$ it holds that : 
\begin{equation}
    \mathbf{H}(\hat{\mathbf{S}})\mathbf{P}^{\top}\mathbf{x} = \mathbf{P}^{\top} \mathbf{H(S)x}
\end{equation}
\end{theorem}
We direct the reader to \cite{gama2019stability} for the proof of Theorem \ref{theorem:permutationequivariance}. The implication of Theorem \ref{theorem:permutationequivariance} is that if the graph exhibits several nodes that have the same graph neighborhoods, then the graph convolution
filter can be translated to every other node with the same neighborhood. We use this key property to bring down the number of training episodes one would need to learn over. In this work in order to keep the topology constant, we assume that the number nearest neighbors each robot is connected to, remains fixed during training and inference.
\section{Graph Policy Gradients for Unlabelled Motion Planning}
The unlabelled motion planning problem outlined in Sec \ref{sec:problem_formulation} can be recast as a reinforcement learning problem. The constraints in Eq. \ref{eq:collision_avoidance} and Eq. \ref{eq:stopping} can be captured by a centralized reward scheme that yield positive rewards only if all goals are covered by robots and negative or zero reward in case of collisions or otherwise. 
\begin{equation}
\label{eq:rewardstruc}
    r(t) =
    \begin{cases}
    \alpha &\text{if }  \phi(t)^\top\phi(t) = \textbf{I}_{\mathbf{N}}\\
   -\beta, &\text{if }  \text{any collisions}\\
   0  &\text{otherwise}
    \end{cases}
\end{equation}
where $\alpha$ and $\beta$ are scalar positive values. Each robot receives this centralized reward. This reward scheme is independent of any constraints on obstacles, dynamics or other constraints on the robots. Let $\Pi=[\pi_1,\ldots,\pi_{\mathbf{N}}]$. Then an approximate solution for \textbf{Problem \ref{prob:prob1}} can be computed
by optimizing for the following loss function: 
\begin{equation}
J = \sum_{n=1}^{\mathbf{N}} \max_{\theta}  \mathbb{E}_{\Pi}\bigg[\sum_{t}^T r_t\bigg]  \enspace
\end{equation}
where $\theta$ represents the parametrization for $\Pi$. In practice we optimize for a $\gamma$ discounted  reward function \cite{sutton1998reinforcement}.

We use graph convolutions to learn filters that aggregate information locally at each robot. $L$ layers of the GCN are used to aggregate information from neighbors $L$ hops away as given in Eqn \ref{eq:finalapproxform}. At every time step $t$, the input to the first layer $z^{0}$ is the vector robot states stacked together, i.e  $z^{0}=\textbf{x}_t = [\mathbf{x}_1,\ldots,\mathbf{x}_{\mathbf{N}}]$ The output at the final layer at any time $t$, $z^{L} =\Pi = [\pi_1,\ldots,\pi_{\mathbf{N}}]$ represents the independent policies for the robots from which actions are sampled.

During training, the weights of the graph filters or the GCN $\theta$, are randomly initialized. Constant graph topology is ensured by fixing the graph at start time. This graph is not varied as the policy is rolled out since the resulting graph convolution yields an equivalent result according to theorem \ref{theorem:permutationequivariance}. 
Each robot rolls out a trajectory $\tau=(\mathbf{x_0},\mathbf{a_0},\ldots,\mathbf{x_T},\mathbf{a_T})$ and collects a reward. It is important to note that this reward is the same for all the robots. The policies for the robots are assumed to be independent and as a consequence the policy gradient $\nabla_{\theta}J$  can be computed directly and is given as: 
\begin{equation}
\label{eq:policygradient}\begin{split}
    \mathbb{E}_{\tau \sim (\pi_1,\ldots,\pi_{\mathbf{N}})}\Bigg[\Big(\sum_{t=1}^T\nabla_{\theta} \log[\pi_1(.)\ldots \pi_{\mathbf{N}}(.)]\Big) \Big(\sum_{t=1}^T r_t \Big) \Bigg]
    \end{split}
\end{equation}
To achieve results that are scalable, we train the filters with a smaller number of robots. This has the additional benefit of reducing the number of episodes required for the policy to converge. During inference, we test with a larger number of robots and execute the learned graph filter over all the robots. Since the filters learn only local information, they are invariant to the number of filters. This is analogous to CNNs where once a filter's weights have been learned, local features can be extracted from an image of any size by simply sliding the filter all over the image. It is important to note that the graph filter only needs centralized information during training. During testing, this solution is completely decentralized as each robot only has a copy of the trained filter and uses it to compute policies that enable all goals to be covered. 
We call this algorithm Graph Policy Gradients (GPG). In the next section, we demonstrate the use of these graph convolutions to learn meaningful policies for different versions of the unlabelled motion planning problem. 
\section{Experiments}
To test the efficacy of GPG on the unlabelled motion planning problem, we setup a few experiments in simulation. To design a reward function that forces robots to cover all goals, we pick the following scheme. For each goal we compute the distance to its nearest robot. The maximum valued distance among these distances is multiplied with $-1$ and added to the reward. We denote this as $r_G(t)$ and it represents the part of the reward function that forces all goals to be covered. However, this does not account for collisions. In order to account for collisions, we compute distance between all robots and add a negative scalar to the reward if the distance between any pair of robots is less than a designed threshold. This part of the reward function is denoted as $r_R(t)$. In the case that the environment is also populated with obstacles, we follow a similar scheme and denote this part of the reward as $r_O(t)$. The overall reward at any time $t$ is then given as a weighted sum of these rewards
\begin{equation}
    r(t) = w_1r_G(t) + w_2r_R(t) + w_3 r_O(t) 
\end{equation} 
where the weights $w_1, w_2$ and $w_3$ are balance the goal following, robot-robot collision avoidance and robot-obstacle collision avoidance. To test GPG, we establish four main experiments. \textbf{1)} Unlabelled motion planning with three, five and  ten robots where robots obey point mass dynamics. \textbf{2)} In the second experiment, GPG is tested on three, five and ten robots but here robots obey single integrator dynamics. \textbf{3)} Here, the robots follow single integrator dynamics and additionally the environment is populated with disk shaped obstacles. \textbf{4)} In this experiment, the performance of GPG is tested against a model based provably optimal centralized solution for the unlabelled motion planning problem. We demonstrate empirically that the performance of GPG is almost always within a small margin of that of the model based method but comes with the additional advantage of being decentralized.   
Closest to our work is that of \cite{khan2019learning} where the robot policies are parametrized by fully connected networks. Thus, to establish relevant baselines, we compare GPG with Vanilla Policy Gradients (VPG) where the policies for the robots are parameterized by fully connected networks (FCNs). Apart from the choice of policy parametrization there are no other significant differences between GPG and VPG. 
\subsection{Experimental Details}
\begin{figure*}[t]
    \centering
    \includegraphics[width=\textwidth]{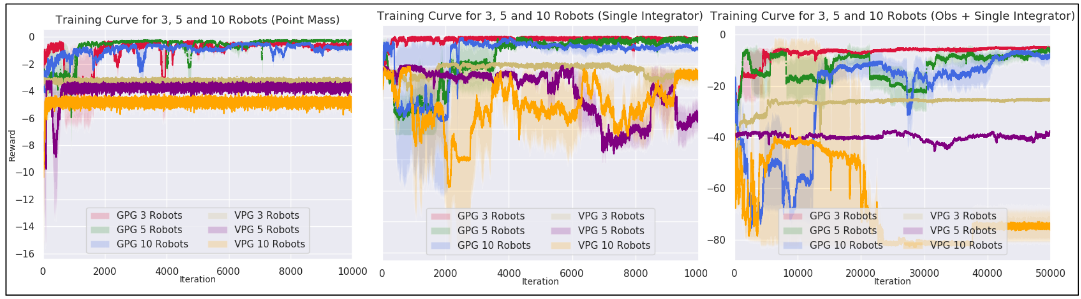}
    \caption{\textbf{Training Curves for 3, 5 and 10 robots} Policies trained by GPG are able to converge on experiments with point mass robots, experiments where robots follow single integrator dynamics and are velocity controlled as well as experiments when disk shaped obstacles are present in the environment. Here iteration refers to episodes. Darker line represents mean and shaded line represents variance.}   
    \label{fig:rew_curves}
\end{figure*}
\begin{figure*}[b]
  \centering
  \includegraphics[width=\textwidth]{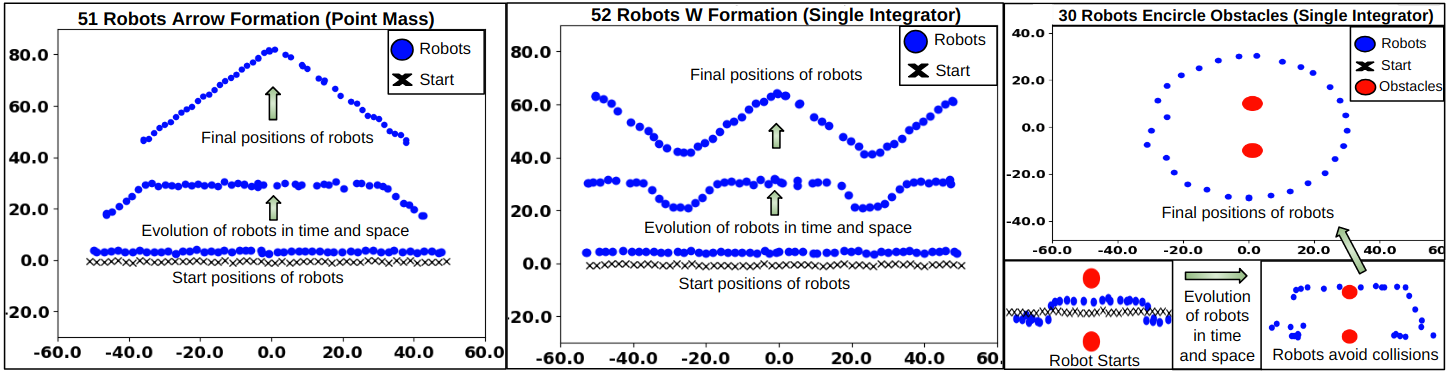}
  \caption{\textbf{Transferring Learned GPG Filters for Large Scale Unlabelled Motion Planning.} (Left) A small number of robots are trained to cover goals and follow point mass dynamics. During testing the number of robots as well as the distance of the goals from the start positions are much greater than those seen during training. (Center) A similar experiment is performed but now robots have single integrator dynamics. (Right) In this experiment in addition to single integrator dynamics, the environment also has obstacles that robots must avoid.\label{fig:formationfig}}
\end{figure*}
For GPG, we setup a L-layer GCN depending on the experiment. For experiments involving 3 and 5 robots with point mass experiments  we find a 2 layer GCN, i,e a GCN that aggregates information from neighbors that are at most 2 hops away to be adequate. For experiments with 10 robots we find that GCNs with 4 layers work the best. For the baseline VPG experiments, we experiment with 2-4 layers of FCNs. The maximum episode length is $200$ steps and the discount factor $\gamma= 0.95$. In experiments with 3 robots, each robot senses 2 nearest goals and 1 nearest robot. In experiments with 5 and more  robots, robots sense 2 nearest goals and 2 nearest robots. The graph $\mathcal{G}$ connects each robot to its $1$,$2$ and $3$ nearest neighbors in experiments with $3,5$ and $10$ robots respectively. 

\subsection{Experimental Results - Training}
The behavior of GPG v/s VPG during training can be observed from Fig. \ref{fig:rew_curves}. We observe that the in all cases GPG is able to produce policies that converge close to the maximum possible reward (in all three cases maximum possible reward is zero). When compared to the convergence plots of \cite{khan2019learning} who first proposed use of RL for the unlabelled motion planning problem, this represents a large improvement on just training.
It can also be observed that GPG converges when robot dynamics are changed or obstacles are added to the environment. While this is not necessarily the optimal solution to the unlabelled motion problem, it is an approximate solution to the unlabelled motion planning problem. The fully connected network policies represented by VPG in Fig. \ref{fig:rew_curves} fails to converge even on the simplest experiments. 

\subsection{Experimental Results - Inference}
The previous section shows the feasibility of GPG as a method for training a large swarm of robots to approximately solve the unlabelled motion planning problem. However, training a large number of robots is still a bottleneck due to the randomness in the system. Training 10 robots with simple dynamics on a state of the art NVIDIA 2080 Ti GPU with a 30 thread processor needs several hours (7-8). We see from our experiments that this time only grows exponentially as we increase the number of robots. 
Thus, to overcome this hurdle and to truly achieve large scale solutions for the unlabelled motion planning problem, we hypothesize that since the graph filters learned by GPG only operate on local information, in a larger swarm one can simply slide the same graph filter everywhere in the swarm to compute policies for all robots without any extra training. Intuitively, this can be attributed to the fact that while the topology of the graph does not change from training time to inference time, the size of the filter remains the same. As stated before, this is akin to sliding a CNN filter on a larger image to extract local features after training it on small images.  
To demonstrate the effect of GPG during inference time, we setup three simple experiments where we distribute goals along interesting formations. As described earlier, each robot only sees a certain number of closest goals, closest robots and if present closest obstacles. Our results can be seen in Fig. \ref{fig:formationfig}. The policies in Fig. \ref{fig:formationfig} (Left) and Fig. \ref{fig:formationfig} (Center) are produced by transferring policies trained to utilize information from 3-hop neighbors. In Fig. \ref{fig:formationfig} the policies are transferred after being trained to utilize information from 5-hop neighbors. Consider the formation shown in Fig \ref{fig:formationfig} (Left). Here each robot receives information about 3 of its nearest goals and these nearest goals overlap with its neighbors. Further, since the goals are very far away and robots are initialized close to each other, a robot and its neighbor receives almost identical information. In such a scenario the robots must communicate with each other and ensure that they each pick a control action such that they do not collide into each other and at the end of their trajectories, all goals must be covered. The GPG filters learn this local coordination and can be extended to every robot in the swarm.

Thus, with these results it can be concluded that GPG is capable of learning solutions for the unlabelled motion planning problem that scale to a large number of robots. 
\subsection{Comparison to Centralized Model Based Methods}
\begin{figure}[b!]
  \centering
  \includegraphics[scale=0.25]{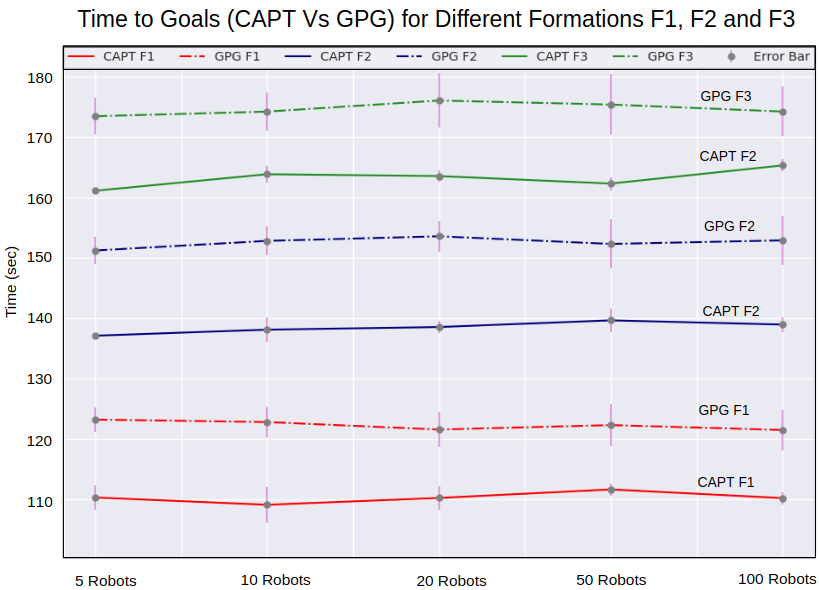}
  \caption{\textbf{Time to Goals (Capt vs GPG)}. Time taken by Capt to cover all goals v/s time taken by GPG to cover all goals when robots follow velocity controls. The goals are arranged in different formations F1, F2 and F3 and differ from each other in terms of distance from the starting positions of the robots (on average goals in F3 > F2 > F1). \label{fig:timefig}}
\end{figure}
To further quantify the performance of GPG, we compare with a model based approach that uses centralized information, called concurrent assignment and planning (CAPT), proposed in \cite{turpin2014capt}. The CAPT algorithm is a probably correct and complete algorithm but needs centralized information. However, when used in an obstacle free environment, it guarantees collision free trajectories. We direct the reader to \cite{turpin2014capt} for more details about the CAPT algorithm. We set up three different formations F1, F2 and F3 similar to that in Fig \ref{fig:formationfig} (Center). On average the goals in F3 are further away from the starting positions of the robots than those in F2 and those in F2 are further away from goals in F1. In this work, we treat CAPT as the oracle and look to compare how well GPG performs when compared to this oracle. We use time to goals as a metric to evaluate GPG against this centralized oracle. Our results can be seen in Fig. \ref{fig:timefig}. 

The key takeaway from this experiment is that decentralized inference using GPG, performs always within an $\epsilon$ margin (approximately 12-15 seconds) of the optimal solution and this margin remains more or less constant even if the goals are further away and if the number of robots are increased. Thus, from this we empirically demonstrate that GPG trades some measure of optimality in exchange for decentralized behavior and this trade-off remains more or less constant even as the number of robots are increased. Hence, with these experiments we conclude that GPG offers a viable solution for Problem \ref{prob:prob1} and is in fact scalable to many robots and further, is very close in performance to the provably optimal centralized solution.

\section{Acknowledgements}
We gratefully acknowledge support from Semiconductor Research Corporation (SRC) and DARPA, ARL DCIST CRAW911NF-17-2-0181, and the Intel Science and Technology Center for Wireless Autonomous Systems (ISTC-WAS) and the Intel Student Ambassador Program.

\section{Conclusion}
In this paper, we look to achieve scalable solutions for the full unlabelled motion planning problem. In the recent past RL approaches have been proposed to compute an approximate solution but these do not scale. In this work, we propose connecting the robots with a naive graph and utilize this graph structure to generate policies from local information by employing GCNs. We show that these policies can be transferred over to a larger number of robots and the solution computed is close to the solution computed by a centralized $\textit{oracle}$. One of the caveats of this paper is the problem of guaranteed collision free trajectories. It might even be possible to add a safety projection set such as that in \cite{khaniros} in order to guarantee collision free trajectories. We leave this for future work.
\addtolength{\textheight}{-12cm}   





\bibliographystyle{IEEEtran}
\bibliography{root}

\end{document}